\documentclass[letterpaper,10pt,conference]{ieeeconf}
\IEEEoverridecommandlockouts

\usepackage{amssymb}
\usepackage{booktabs}
\usepackage{color}
\usepackage[backend=biber,
            url=false,
            isbn=false,
            doi=false,
            backref=false,
            style=ieee,
            natbib=true,
            mincitenames=1,
            maxcitenames=1,
            citestyle=numeric-comp,
            sorting=none,
            block=none]{biblatex}
\renewcommand{\bibfont}{\small}
\usepackage{graphicx}
\let\labelindent\relax
\usepackage{enumitem}
\usepackage{amsmath}
\usepackage{wrapfig}
\usepackage{caption}
\usepackage{subcaption}
\usepackage{tabularx,array}
\usepackage{listings}
\usepackage{xcolor}
\usepackage{tcolorbox}

\usepackage[colorlinks=true,
            linkcolor=blue,
            citecolor=blue,
            urlcolor=blue]{hyperref}
\usepackage[nameinlink, capitalize]{cleveref} 

\usepackage{siunitx}
\usepackage{caption}
\definecolor{lightgray}{gray}{0.95}

\lstdefinelanguage{yaml}{
  morekeywords={true,false,null,yes,no},
  sensitive=false,
  morecomment=[l]{\#},
  morestring=[b]",
  morestring=[b]'
}

\makeatother
\title{\LARGE \bf
 AURA: Autonomous Upskilling with Retrieval-Augmented Agents\\
}
\author{
Alvin Zhu$^{1*}$, Yusuke Tanaka$^{2*}$, Andrew Goldberg$^{3}$, Dennis Hong$^{2}$%
}

\begin{document}
\twocolumn[{%
\renewcommand\twocolumn[1][]{#1}%
    \maketitle
    \begin{center}
        \centering
        \includegraphics[width=0.95\textwidth,trim={0cm 0cm 0cm 0cm},clip]{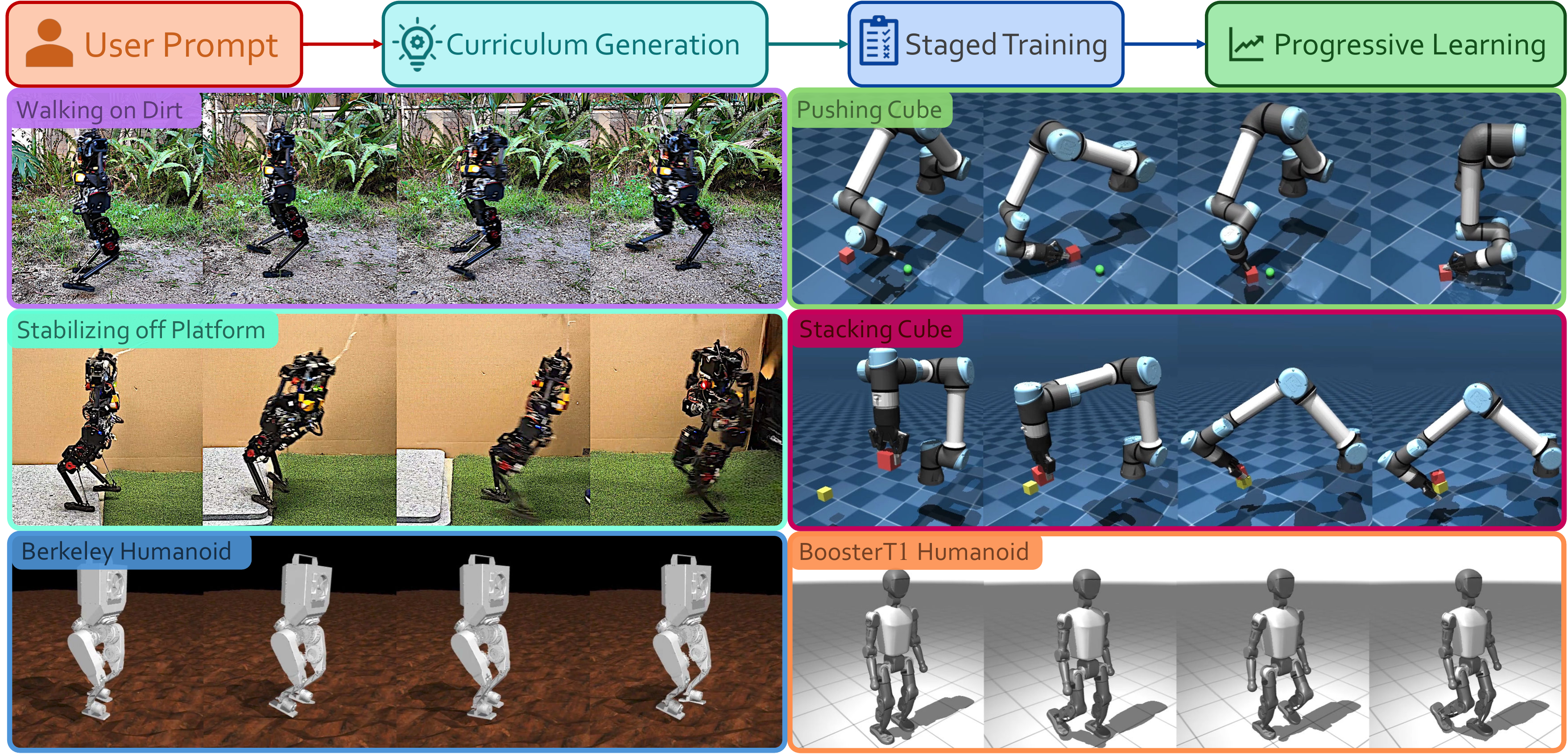} %
    \captionof{figure}{
    AURA-trained policies deployed successfully on custom humanoid hardware and in simulation for locomotion and manipulation tasks.
    }
    \label{fig:figure1}
    \end{center}%
    }]
    \footnotetext[1]{A. Zhu is with Department of Computer Science and Electrical Engineering, $^{2}$Y. Tanaka and D. Hong are with Department of Mechanical and Aerospace Engineering, UCLA, Los Angeles, CA, USA. $^{3}$A. Goldberg is with Department Electrical Engineering and Computer Science, UC Berkeley, Berkeley, CA, USA. \{alvin.zhu, yusuketanaka, dennishong\}@g.ucla.edu. apgoldberg@berkeley.edu.
$^*$ denotes equal contribution.}%
\thispagestyle{empty}
\pagestyle{empty}

\begingroup


\begin{abstract}
Designing reinforcement learning curricula for agile robots traditionally requires extensive manual tuning of reward functions, environment randomizations, and training configurations. We introduce \textbf{AURA (Autonomous Upskilling with Retrieval-Augmented Agents)}, a schema-validated curriculum reinforcement learning (RL) framework that leverages Large Language Models (LLMs) as autonomous designers of multi-stage curricula. AURA transforms user prompts into YAML workflows that encode full reward functions, domain randomization strategies, and training configurations. All files are statically validated before any GPU time is used, ensuring efficient and reliable execution. A retrieval-augmented feedback loop allows specialized LLM agents to design, execute, and refine curriculum stages based on prior training results stored in a vector database, enabling continual improvement over time. Quantitative experiments show that AURA consistently outperforms LLM-guided baselines in generation success rate, humanoid locomotion, and manipulation tasks. Ablation studies highlight the importance of schema validation and retrieval for curriculum quality. AURA successfully trains \textbf{end-to-end} policies directly from user prompts and deploys them \textbf{zero-shot} on a custom humanoid robot in multiple environments
—capabilities that did not exist previously with manually designed controllers. By abstracting the complexity of curriculum design, AURA enables scalable and adaptive policy learning pipelines that would be complex to construct by hand. 
\href{https://aura-research.org/}{aura-research.org}
\end{abstract}




	

\section{Introduction}
Curriculum reinforcement learning (RL) \cite{narvekar2020curriculum, Li2024RLbipedal, anymalparkourlearningagile, fourierlatentdynamics} enables robots to master complex skills by decomposing tasks into progressively harder subtasks \cite{rl_curriculum}. This approach has proven to be effective in domains like agile locomotion \cite{sciencesoccer, real_world_digit, massive_rl, learningtorquecontrolquadrupedal}, where single-stage learning can struggle with sparse rewards or high-dimensional exploration spaces. By splitting the learning process, agents are guided through increasingly challenging environments, improving sample efficiency and convergence.

However, designing effective multi-stage curricula remains a significant bottleneck \cite{jang2018combining, nguyen2021robots}. 
Multi-stage curricula require coordinated changes across reward shaping, randomization, simulation fidelity, and optimization; as the number of stages increase, designing transitions, verifying stability, and tuning parameters scales combinatorially (the “curse of dimensionality” \cite{li2025causally}) and becomes brittle to human error and heuristic bias \cite{sayar2024diffusionCurriculum, soviany2022curriculum}.
Moreover, failures at any stage often derail the entire training process, making automation critical for scaling curriculum RL.

Large language models (LLMs) offer a compelling alternative to optimization-based approaches \cite{hu2020learning}. Their high-level reasoning and generalization capabilities have demonstrated success in robotic planning, perception, and code generation \cite{brohan2022rt, ahn2022can, liang2023code, sun2024prompt, mower2024ros, mon2025embodied}. 
While LLMs are not directly deployable on real-time control loops due to latency constraints \cite{team2025gemini}, LLM's have shown potential in \emph{designing} training schemes for real-time robot control policies \cite{ma2023eureka, ryu2024curricullm, ma2024dreureka}. However, current LLM-based RL pipelines inefficiently use computation on launching parallel environments to deal with malformed generations and fail to learn from experience across tasks.


What is missing is a principled framework for transforming high-level prompts into reliable, executable RL training pipelines that improve with experience. Such a system should (1) ensure syntactic and semantic validity before execution, (2) use past experience to optimize future performance, and (3) modularize the pipeline into verifiable components \cite{Erak_2024}. Without these capabilities, LLMs are limited to more fragile, trial-and-error reward design rather than scalable curriculum generation.

\begin{figure*}
    \centering
    \includegraphics[width=0.9\linewidth]{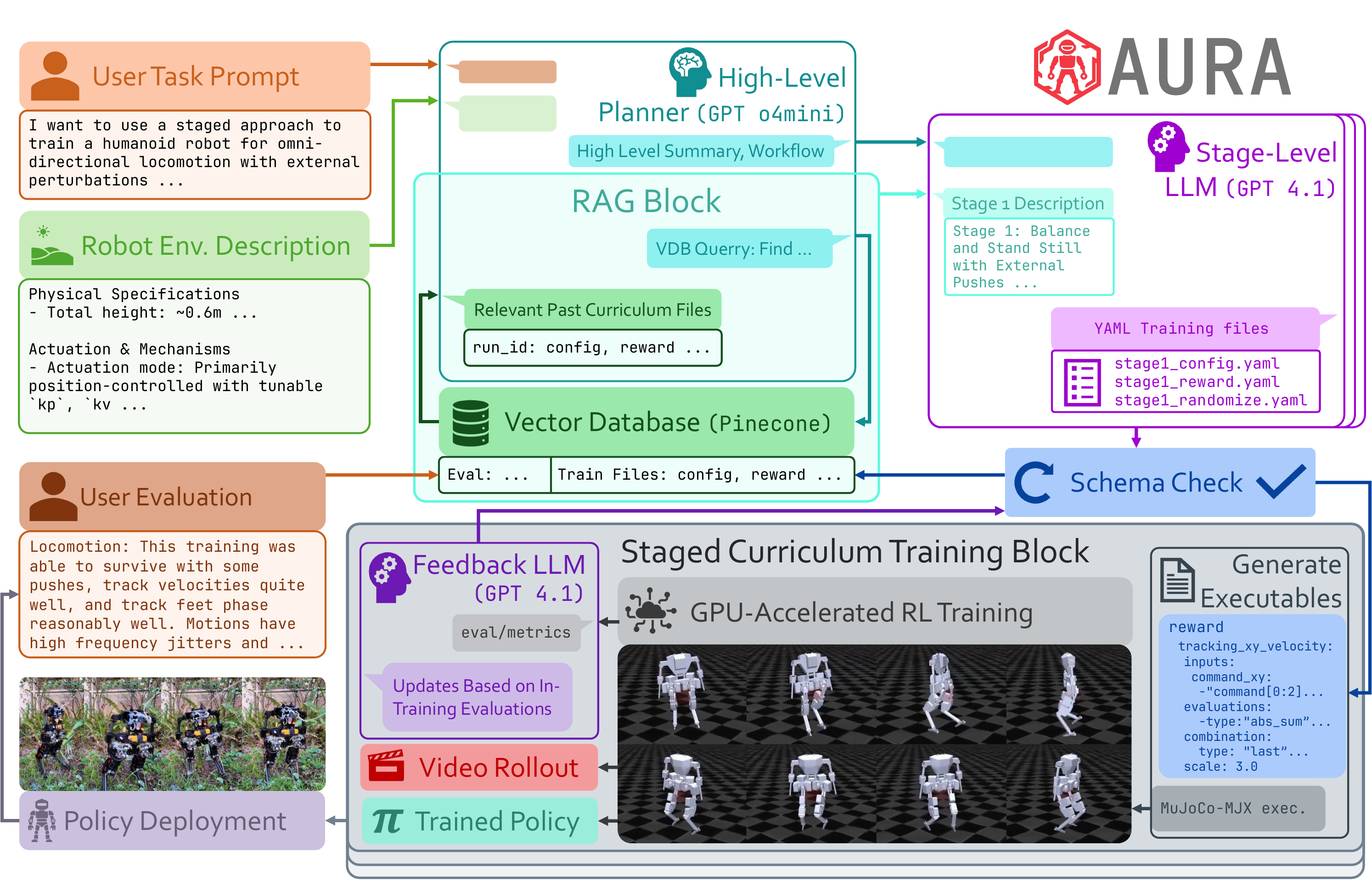}
    \caption{An overview of the AURA curriculum generation and policy training framework.}
    \label{fig:AURA-framework}
\end{figure*}

\textbf{AURA} (\emph{Autonomous Upskilling with Retrieval-Augmented Agents}) addresses these bottlenecks with three key ideas.  
(i) A \emph{typed YAML schema} captures curricula, reward functions, randomization, and training hyperparameters; structured LLM outputs are statically validated before a single GPU cycle is spent.  
(ii) A \emph{team of specialized LLM agents} works in collaboration to enable modular, schema-compliant curriculum generation.
(iii) A \emph{Retrieval-augmented generation (RAG)} module with a \emph{vector database (VDB)} stores prior task specifications, curricula, and rollout evaluations. This database supports experience-informed generation by enabling agents to condition on successful prior workflows, improving both accuracy and generalization.

We evaluate AURA on a suite of humanoid locomotion and deploy the resulting policies zero-shot on a custom, kid-sized humanoid. Compared to curriculum RL baselines and recent LLM-crafted reward pipelines, AURA is capable of generating context-rich, high-dimensional curricula that produce successful zero-shot policies deployable on real humanoid hardware. Our contributions are:

\setlist[enumerate]{leftmargin=12pt}
\begin{enumerate}
    \item \textbf{AURA}: A fully agentic, retrieval-augmented framework that turns a natural-language prompt into a hardware-deployable controller policy.
    \item \textbf{Curriculum compiler, schema-validation}:  
        A typed YAML schema that provides an LLM-friendly interface for defining reward terms, domain randomizations, and training configurations. Static validation of the YAMLs enables descriptive error messages for generation retries, without wasting compute on failed environment launches.
    \item \textbf{Experience-aware, self-improving, curriculum generation agent}: 
        Specialized LLM agents query a vector database of past runs and evaluations, select relevant training files, generate high-level curricula, and refine them into executable multi-stage specifications. This feedback-driven loop promotes curriculum quality and training stability over many iterations.
    \item \textbf{Zero-shot deployment}: Experiments showing AURA’s policies transfer zero-shot to a custom kid-sized humanoid, Berkeley Humanoid, Booster T1, \cite{berkeleyhumanoid}, and manipulators, showing end-to-end prompt-to-policy ability.
\end{enumerate}

\section{Related Work}
\subsection{Large Language Models in Robotics}
Early efforts to link natural language to robot control and task-planning frameworks translate natural-language commands into symbolic action graphs, as demonstrated by SayCan and Code-as-Policies \cite{ahn2022can, liang2023code}.
Recent Vision–Language–Action paradigms, such as RT-1/2, $\pi0$, Octo, and OTTER, \cite{brohan2022rt,zitkovich2023rt, pi0, octo, otter}, demonstrate that pretraining on large-scale robot data yields policies that generalize across many robotic skills.
Further work has extended this to cross-domain generalization, deformable object manipulation, and long-horizon household tasks, exemplified by Gemini Robotics and embodied LLMs \cite{mon2025embodied, team2025gemini}. 
Other lines of work have explored LLM-driven planning and language-augmented reasoning for robotic control pipelines \cite{sun2024prompt, mower2024ros, mon2025embodied, llm_robotics_survey, ai_review_humaniod, llm_robot_task_plan, rho2024language}.  
Despite their semantic flexibility, these methods continue to rely on hand-tuned, low-level controllers and standard actuation assumptions, with LLMs typically external to the closed control loop \cite{llm_robotics_survey, ai_review_humaniod}. RoboGen addresses these limitations through decomposing long-horizon tasks into skills, which are tuned or trained by LLMs \cite{auerbach2014robogen}.

\subsection{Curriculum Reinforcement Learning}
Reinforcement learning often struggles with sparse rewards and long‐horizon problems, where meaningful feedback is rare or delayed over many timesteps \cite{discover}. Curriculum RL addresses sparse-reward and long-horizon problems by exposing agents to a progression of gradually more difficult tasks \cite{narvekar2020curriculum,rl_curriculum,soviany2022curriculum}.
Surveys and benchmarking studies note that practical curriculum design remains heuristic and sensitive to context-specific particulars \cite{soviany2022curriculum}.
Reverse-curriculum approaches \cite{reverse_curriculum_rl, tao2024reverse} employ a staged training process that progressively increases the task's difficulty from previous successful demonstrations, thereby enhancing exploration and sample efficiency.
Domain-randomization and curriculum-based RL have enabled real-world legged locomotion and manipulation in challenging environments \cite{Li2024RLbipedal, anymalparkourlearningagile, fourierlatentdynamics, sciencesoccer, real_world_digit,  massive_rl, learningtorquecontrolquadrupedal, zhuang2024humanoidparkourlearning, quadsoccershoots, dribblebot,nimbro_2022, adu2023exploring}. 

\subsection{LLM-guided RL and Curricula}
Emerging systems such as CurricuLLM \cite{ryu2024curricullm} and Eureka \cite{ma2023eureka, liang2024eurekaverse} utilize LLMs to generate reward functions for RL training. Eureka uses evolutionary search and prompts LLMs in parallel to directly generate reward code for various tasks in simulation. DrEureka extends Eureka to generate domain randomizations and consider hardware safety, demonstrating real-world locomotion tasks on a Unitree Go1 quadruped robot \cite{ma2024dreureka}. CurricuLLM \cite{ryu2024curricullm} similarly uses evolutionary search and directly generates code, but focuses on curriculum generation and humanoid locomotion, showing real-world success on the Berkeley Humanoid \cite{berkeleyhumanoid}.

These frameworks require sampling a large number of training runs to obtain a viable or high-performing policy, relying on LLM stochasticity and brute-force search. RAG and VDBs are approaches to enhance the LLM knowledge beyond what the LLM has been trained on \cite{liu2025hm}, which AURA utilizes to construct effective curricula. 

\textit{Positioning of this work}: AURA unifies curriculum generation, domain randomization, and training configuration within a schema-validated, retrieval-augmented loop that learns from prior runs. This yields consistent, deployable specifications with higher first-attempt launch rates and removes the need for post hoc sampling or manual curation.


\section{Methods}
\label{sec:methods}

\subsection{Problem Setup and Curriculum Formalism}
We define a curriculum $\mathcal{C}$ as an ordered sequence of $K$ training stages,
$\mathcal{C}=\{\xi_1,\dots,\xi_K\}$, where each stage $\xi_k=\langle \Phi_k,\rho_k,\Theta_k,\kappa_k\rangle$ is comprised of a reward $\Phi_k$, domain-randomization distributions $\rho_k$ over environment
parameters $\psi$, a training configuration $\Theta_k$, and a promotion criterion $\kappa_k(\pi)$ that governs advancement. The curriculum stages are modeled as an MDP
$\mathcal{M}=\langle \mathcal{S},\mathcal{A},\mathcal{T}(\cdot\mid s,a;\psi),R_{\psi},\gamma\rangle$, \cite{sutton1998reinforcement, puterman2014markov}
with $\psi$ denoting randomized physics properties \cite{tobin2017domainrandomization, muratore2021data}.
AURA generates $(\Phi_k,\rho_k,\Theta_k)$ for each stage as schema-compliant YAML files that are statically
validated and compiled into an MJX-based RL pipeline \cite{todorov2012mujoco, brax2021github}. 
We use this notation to describe generation, validation, training, and iteration.

\subsection{Inputs to Curriculum Generation}
To instantiate a curriculum from a prompt, AURA consumes four inputs:
(i) a natural language task description specifying the desired behavior;
(ii) a structured robot specification: joint names, semantic labels (e.g., ``left foot contact''), available
sensors, and actuator limits;
(iii) an MJCF simulation model encoding robot kinematics, dynamics, and contact geometry; and
(iv) a Python MJX environment defining observations, actions, physics parameters, and episode termination conditions.

\subsection{Retrieval-Augmented High-Level Planning}
AURA begins by leveraging LLMs and a vector database of prior curricula and outcomes to generate a high-level plan. 
The user-provided task description and robot specification are first passed into a \emph{query LLM}, which synthesizes them into a structured retrieval query. 
This query is then encoded with OpenAI’s \texttt{text-embedding-3-large} \cite{openai_text_embedding_v3} and used to perform cosine-similarity search over the VDB (Pinecone). 
From the top-$3$ most relevant prior curricula, the database returns reward, randomization, and training files, along with user feedback.
A \emph{selector LLM} then chooses the single most relevant example from this set, based on both user feedback and curricula relevance to the current task; this example is included in the high-level planner's context for generating the new curricula.

Conditioned on this selected example and the user task specification, a high-level planner decides the number of stages and produces a natural-language plan for each one. 
The plan describes ideas for reward components, domain randomizations, and stage-specific training hyperparameters (e.g., terrain type, disturbance level, PPO hyperparameters, etc.). 

 The VDB enables two modes of AURA. (1) AURA Blind where the VDB is initialized as empty for the first iteration and AURA must design a reward from scratch, and (2) AURA Tune where the VDB is initialized with human-expert rewards from a related task and must \emph{tune} that reward for the current domain.




\subsection{Stage-Level Specification and YAML Generation}
Each stage plan from the high-level planner is given to an independent stage-level generator which translates the stage plan into structured YAML files. The stage-level generator takes as input:
(i) the user task description; (ii) the natural-language stage description produced by the high-level planner;
(iii) the available reward function library; (iv) state variables and sensor signals exposed by the robot environment;
(v) the selected prior curriculum example retrieved from the VDB (chosen from the top-$3$ retrieval set);
(vi) context from previously generated stages in the current workflow; and (vii) the reward/schema format specifications.

Each stage description $\xi_k$ is translated into three human-readable, schema-compliant YAML files: \emph{Reward YAML} $(\Phi_k)$ defines the differentiable terms, coefficients, and aggregation logic with explicit sensor/state dependencies; \emph{Randomization YAML} $(\rho_k)$ defines the target parameters, distributions, and activation conditions; and \emph{Training Config YAML} $(\Theta_k)$ defines the PPO hyperparameters, episode budgets, checkpointing cadence, normalizations, etc.. These files provide the abstract specifications needed for execution without requiring direct generation of JAX code.


\subsection{Schema Validation and Compilation}
Directly emitting JAX/MJX code from LLMs can be brittle in practice. AURA compiles only after static validation
against typed schemas governing the generated workflow, reward, randomization, and training files. Validation enforces:
(i) type conformance (\texttt{int}, \texttt{float}, \texttt{bool}, \texttt{vector}, etc.);
(ii) structural schema compliance; (iii) reference integrity (all state variables and sensors referenced in $\Phi_k$
exist in the MJX environment); and (iv) mathematical well-formedness of reward expressions. Expressions are built from
a function registry (e.g., \texttt{fn.NORM\_L2} producing JAX arrays) and must compile symbolically under MJX.
A generation retry is triggered whenever schema validation detects an error. The schema validation generates a descriptive error message that includes the type of error, content of what caused the error, and sometimes a recommendation for how the error can be fixed. This error message, along with the previously generated (but invalid) YAML files, are fed back into the next attempt to guide corrections and avoid repeating mistakes; we set a maximum of five retries per stage. Once validated, YAMLs are compiled into MJX code.

\subsection{Staged RL Training Loop}
Compiled stages are trained with PPO across parallel simulation environments, each running for the episode limit specified in $\Theta_k$. Policies from one stage initialize the next, enabling progressive skill learning.

\subsection{Feedback and Iteration}
Following each stage, an automated feedback module analyzes rollouts and training signals under a task-specific rubric
(e.g., success rate, stability and energy, reward-component attribution). Its recommendations may adjust reward terms,
randomizations, or hyperparameters for \emph{subsequent stages within the same iteration} if the curriculum training has not yet completed all $K$ stages.

An \emph{iteration} of our framework is defined as the complete execution of:
(i) VDB-backed retrieval and high-level planning; (ii) YAML generation and static validation for all stages;
(iii) PPO training across $K$ stages; with (iv) automated feedback analysis between each stage. After the final trained policy is deployed in simulation or on hardware, the user will evaluate the policy either based on the simulation video rollout or hardware performance, then provide \emph{user feedback}: 2--4 sentences in natural language judging the policy's behavior and quality based on the video rollout (eg. "the policy survives well, but could track linear velocity better"). 
User feedback is attached to the files of the just-trained curriculum and inserted into the VDB alongside the YAMLs,
metrics, and rollouts. Any training that incorporates user feedback begins a \emph{new} iteration $(t{+}1)$, which again
starts from VDB retrieval conditioned on the now-augmented database. Human input therefore does not mutate an ongoing
run; it shapes the next planning phase via retrieval.

\subsection{Modular LLM Collaboration}
\textit{Hallucination mitigation}:
AURA limits LLM hallucinations through three mechanisms: (i) \emph{retrieval grounding}—the planner is context-seeded by a single example selected from the top-$3$ VDB results; (ii) \emph{schema constraints}—the model emits only typed YAML using a restricted operator registry (e.g., \texttt{fn.NORM\_L2}) rather than arbitrary code; and (iii) \emph{consistency checks}—static validation enforces type/structure and reference integrity to environment signals, followed by MJX compile checks and a capped regenerate-on-error loop (max $5$ retries). 
Together, these mechanisms reject unsupported symbols and incoherent references, improve first-try success, and reduce off-distribution generations.


\begin{figure*}
    \centering
    \includegraphics[width=0.9\linewidth]{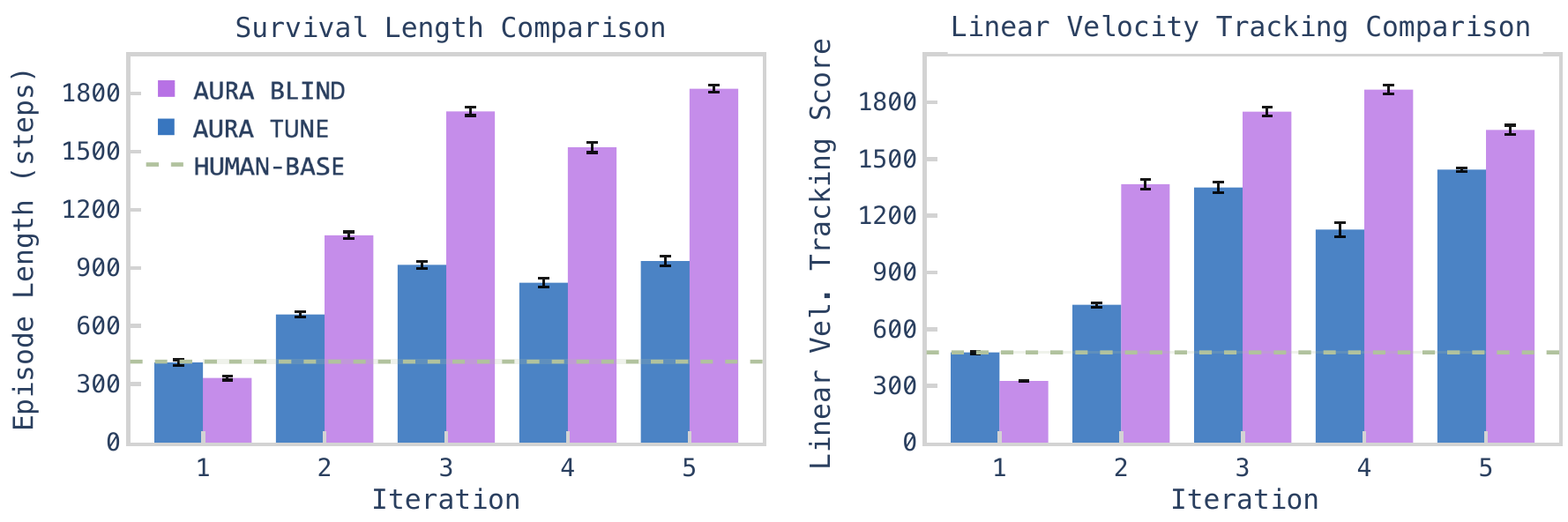}
    \caption{Survival and linear velocity tracking scores across iterations to evaluate locomotion policy quality on a custom humanoid. 
    The plots show the policy quality improvements of AURA over five iterations compared to MuJoCo Playground's expert designed Berkeley Humanoid reward and CuricuLLM's reported results in Isaac Lab. AURA Blind generates rewards from scratch (VDB is initialized as empty) and AURA Tune modifies and improves an existing reward designed for another embodiment (VDB is initialized with MuJoCo Playground's Berkeley Humanoid expert human rewards, domain randomizations, and training configuration).
    }
    \label{fig:training}
\end{figure*}

\section{Experiments}
\subsection{Simulation Environments}
We present simulation experiments on a variety of tasks on three humanoid robots and a robot arm manipulator, all example AURA trained policy rollouts can be seen in \cref{fig:figure1}. 
All AURA experiments are trained in MuJoCo-MJX and are evaluated over 1024 parallel environments on 5 seeds. CurricuLLM simulation results are reported based on their results in Isaac Lab.

\subsubsection{Custom Humanoid}
The custom humanoid locomotion environment exposes the velocity command, joint positions, projected gravity vector, last action, phase, and max foot height as observations and actuator gains, friction coefficients, body mass, COM position, foot contact geometry, initialization configuration, and roughness as domain randomization parameters. The simulation setup was adapted from the model in \cite{bruce_parallel_sim}. AURA is run for five iterations, with each iteration generating a new curriculum. Custom humanoid locomotion is evaluated with AURA Tune, which starts with the Berkeley Humanoid reward and adapts it into a curriculum that works on the custom humanoid hardware, and AURA Blind, which generates a curriculum from scratch. All environments are run at $50$Hz, and the curriculum is limited to $400$M training timesteps.

\begin{table}[t]
\centering

\renewcommand{\arraystretch}{1.4} 
\small
\begin{tabularx}{\columnwidth}{@{}l *{2}{>{\centering\arraybackslash}X}|>{\centering\arraybackslash}X@{}}
\toprule
& \shortstack{Episode Survival\\Length} 
& \shortstack{Linear Vel.\\Tracking} 
& \shortstack{Pushing Cube\\Success \%} \\
\midrule
AURA      & \mbox{\footnotesize $2873 \pm 474.8$} & \mbox{\footnotesize $\textbf{2077} \pm 425.6$} & \mbox{\footnotesize $\textbf{91.95} \pm 0.95$} \\
Playground      & \mbox{\footnotesize $\textbf{2903} \pm 411.1$} & \mbox{\footnotesize $1546 \pm 375.8$} & $0.00 \pm 0.00$** \\
CurricuLLM* & \mbox{\footnotesize $315.4 \pm 73.95$}   & \mbox{\footnotesize $40.46 \pm 10.64$}              & \mbox{\footnotesize $62.50 \pm 16.60$} \\
\bottomrule
\end{tabularx}
\captionsetup{font=footnotesize, labelfont=bf, textfont=normalfont}
\caption{Policy evaluation across metrics. Episode survival length and linear velocity tracking are used to evaluate a velocity command following task on the Berkeley Humanoid.
The success rate of pushing cubes is evaluated on the UR5e enviorment. *CurricuLLM's Berkeley Humanoid and Fetch-Push results are reported in the paper \cite{ryu2024curricullm}. **MuJoCo Playground's Pushing Cube Success is reported using Franka Emika Panda rewards on the UR5e embodiment, which shouldn't be expected to be successful. AURA adapts the Franka expert reward and training configuration into an effective curriculum for the UR5e.}
\label{tab:policy_eval}

\end{table}

\subsubsection{Berkeley Humanoid}
For the Berkeley humanoid task, AURA uses the MuJoCo Playground's Berkeley Humanoid environment.
Berkeley Humanoid's MuJoCo Playground policy is trained as detailed in \cite{mujoco_playground_2025}. The environment is run at 50Hz and both policies train over $300$M steps.

\subsubsection{BoosterT1 Humanoid} For BoosterT1 locomotion, AURA is trained and evaluated in the MuJoCo Playground BoosterT1 simulation environment, is run at 50Hz, and trains for $300$M steps. BoosterT1's MuJoCo Playground policy is trained as detailed in \cite{mujoco_playground_2025}. 

\subsubsection{UR5e Manipulator}
The environment observations and domain randomizations are consistent with MuJoCo Playground's manipulator environment for the Franka Emika Panda (but with a UR5e robot). The environment frequency is $30$Hz, and the curriculum is limited to $300$M training steps. This environment is used for two tasks. In the pushing cube task the robot must push a block to a target position both of which are randomized in a $30$cm by $30$cm region (following CurricuLLM experiments). To be considered a success the center of the cube must be within $3$cm of the target point. The stacking cube task begins with a cube in the gripper and the robot must place the cube on top of another block which is randomized in a $30$cm by $30$cm region. 


\begin{table*}[t]
\centering

\small
\begin{tabular}{@{}lcccccc@{}}
\toprule
 & \textbf{AURA} 
 & AURA w/o schema 
 & AURA w/o VDB 
 & AURA single-agent 
 & CurricuLLM$^{*}$ 
 & Eureka$^{**}$ \\
\midrule
\begin{tabular}[c]{@{}l@{}}Training-launch\\Success Rate\end{tabular}
 & \textbf{99\,\%} & 47\,\% & 38\,\% & 7\,\% & 31\,\% & 12 / 49\,\% \\
\bottomrule
\end{tabular}
\caption{
Training-launch-success-rate comparing AURA and its ablated variants. 
All evaluations are conducted with GPT-4.1, as the original models used in the baselines are deprecated at the time of assessment. 
$^{*}$CurricuLLM is evaluated on generating rewards for Berkeley Humanoid locomotion.  
$^{**}$Eureka's $12\%$ is evaluated on generation for their ANYmal task, which is most similar in complexity to humanoid robot tasks. Eureka's generation success rate across all available embodiments in their examples is $49\%$ with simpler tasks generating more successfully.
}
\label{tab:execution_success}
\vspace{0.25em}

\end{table*}

\subsection{Experimental Setup}
We evaluate two aspects of AURA and baselines:

\emph{(1) Generation Success Rate.} 
In \cref{tab:execution_success}, we report the success rates of AURA, its ablations, CurricuLLM, and Eureka. 
For AURA, we measure \emph{curriculum}-level success. A run is considered successful only if \emph{all} stages (rewards, domain randomizations, and training configs) in the generated curriculum successfully launch GPU-accelerated training. AURA performs a single launch per task without parallel sampling. We also compare AURA with its ablated variants: AURA without a VDB, AURA without schema validation (no retires), and AURA using a single LLM for both high-level planning and YAML generation. Eureka and CurricuLLM launch multiple environments in parallel. We report the fraction of successful environment launches out of all generated rewards. Unlike AURA, success is measured one stage at a time rather than requiring a full curriculum to succeed.

\emph{(2) Training and Deployment Benchmarking.} 
To evaluate policy quality, we run five iterations of curriculum generation and training using AURA, AURA Blind, AURA Tune, CurricuLLM, and a human-designed baseline. Each trained policy is evaluated with $5$ random seeds each across $1024$ randomized environments. Evaluation reward functions, domain perturbations, and task variations are hidden from all methods to simulate realistic zero-shot deployment.
For locomotion, survival episode length is the average number of steps survived without the humanoid falling (max 3000).
Linear velocity tracking score is calculated per episode as:
\[
\mathcal{S}_{\text{lin}}=\sum_{t=1}^T\exp\!\Bigl(-\frac{\lVert \mathbf{v}^{\mathrm{cmd}}_{t}-\mathbf{v}^{\mathrm{loc}}_{t}\rVert_2^{2}}{2(\sigma)^2}\Bigr), \:\sigma = 0.1
\]
Where $v^{cmd}_t$ and $v_t^{loc}$ are the commanded and actual robot velocities at timestamp $t$ and $T$ is the number of timsteamps the policy survives for.

Selected final policies are transferred to hardware for qualitative testing and demonstration.





\subsection{Simulation Experiment Evaluations}
\subsubsection{Custom Humanoid}
\cref{fig:training} summarizes evaluation during simulation deployment across iterations, showing average survival and linear velocity tracking scores. 
Both AURA variants improve noticeably over the human baseline on survival and linear-velocity tracking, which shows that AURA can generate an effective curriculum from scratch and can adapt existing rewards into effective curriculums. AURA Tune and AURA Blind both iteratively improve over five iterations, reasoning about reward signals and user feedback. AURA Blind's curriculum did not generate or use any rewards related to foot swing or foot phase tracking, leading to a more inconsistent and asymmetrical gait, but better simulation performance compared to AURA Tune. The locomotion task prompt does not specify stylistic information about the gait, so this is a reasonable result. AURA Tune, which is initialized with MuJoCo Playground's rewards, kept existing reward terms for phase tracking, maintaining a consistent gait while still improving over the base reward.

\subsubsection{Berkeley Humanoid}
For the Berkeley Humanoid results shown in \cref{tab:policy_eval}, AURA’s episode survival length is similar to the MuJoCo Playground baseline, showing that both survive nearly the full 3000-step horizon. The key difference is in linear velocity tracking, where AURA achieves a significantly better score, reflecting more precise command following. While Playground training schedules $200$M steps of flat terrain then $100$M steps on rough terrain, AURA split training into $100$M on flat terrain with mild perturbations, $100$M on rough terrain with mild perturbations, and $100$M on rough terrain with heavier, more frequent pushes, while progressively expanding the command range up to the 1.0 m/s and 1.0 rad/s caps. AURA also made the velocity following reward more strict and increased reward emphasis on feet-phase and action-rate, resulting in better velocity command following and smoother, more consistent stepping.

\subsubsection{BoosterT1 Humanoid}
Training with the MuJoCo Playground's BoosterT1 reward directly resulted in a policy with jerky motion and small, high-frequency shakes during locomotion. AURA was told this as feedback, which led it to generate new reward terms which penalize the squared differences across consecutive time steps for actions and for velocities, resulting in a visibly smoother policy. Heavier domain randomization also helped improve performance, raising episode length from $2139$ to $2366$ steps and linear-velocity tracking score from $1786$ to $2162$.

\subsubsection{UR5e Manipulator}
For the UR5e cube pushing task, AURA begins with MuJoCo Playground's Franka Emika Panda environment reward which is not sufficient to train a policy in the UR5e environment as shown in \cref{tab:policy_eval}. AURA adapted the reward into a curriculum where the first stage has no positional domain randomization and reshaped the gripper–to-cube reward term to provide a denser reward signal over larger distances, encouraging progress toward the cube rather than only near contact. In the second stage, AURA reintroduced the positional domain randomization and increased the weight of the box to target reward. This led to a final policy which could consistently push the cube to the target location with over a $90$\% success rate. 

For the cube stacking task AURA adapted the push task curriculum, dramatically decreasing the gripper-to-cube reward because the cube starts in the gripper, and emphasizing the cube-to-cube reward term. AURA achieved a mean task success rate of 72.7\%.



\subsection{Real-World Experiments}
\textit{Hardware Setup}:
We use a custom $0.6$ m-tall agile humanoid with 5-DoF legs and 3-DoF arms \cite{bruce}. RL policies run at $50$ Hz with sensors and actuators at $500$ Hz. The end-to-end policy uses no estimators or low-level control beyond servo internal impedance controllers.

\cref{fig:figure1} shows a policy generated by AURA deployed on real hardware in a zero-shot setting. The outdoor locomotion policy shown in the figure is trained by AURA Tune, while the robot traversing a step is trained by AURA Blind. In response to a user prompt requesting outdoor-capable locomotion, AURA automatically generated a four-stage curriculum progressing from flat terrain with low randomization to rough terrain with high randomization, diverse velocity commands, and significant external perturbations.
The resulting policy demonstrated robust zero-shot generalization when deployed outdoors, successfully handling uneven terrain and maintaining consistent gait tracking. The robot achieved walking at $0.18$ $\mathrm{m/s}$ and average gait frequency of $1.91$ $\mathrm{Hz}$. It withstood substantial perturbations, recovering from lateral pushes up to $0.38$ $\mathrm{m/s}$ and from being physically pushed off a $50$ $\mathrm{mm}$ high platform without falling, which surpassed the performance of existing manually tuned controllers \cite{bruce}.

\subsection{Generation Success and Efficiency}
\subsubsection{Curriculum Generation} ~\cref{tab:execution_success} presents the success rates of training runs for AURA, its ablated variants, and baseline methods including CurricuLLM and Eureka. AURA achieves a \textbf{$99\%$} success rate, outperforming all other methods and demonstrating the benefit of schema validation and RAG which allow for consistent generation of curriculums with hundreds of parameters and sometimes over ten unique reward terms. 

\subsubsection{Computation Efficiency}
AURA requires only a \emph{single} training launch per stage, since schema validation and VDB-guided retrieval ensure that nearly all generated curricula are executable on the first training launch. In contrast, Eureka and CurricuLLM each generate multiple training launches per stage—16 and 5, respectively—and retain only the best-performing successful run. As a result, AURA initiates far fewer training runs than either CurricuLLM or Eureka, leading to substantially lower computational cost for multi-stage workflows while still achieving competitive performance.

\subsubsection{Multi-Agent LLM Setup} 
AURA's use of a multi-agent LLM setup—including a high-level curriculum generation agent and dedicated per-stage agents for reward, domain randomization, and configuration—was critical for producing complete and valid curricula. As shown in ~\cref{tab:execution_success}, attempting to generate all stages in a single step is highly inconsistent. By decomposing the task into high-level planning and individual stage generation AURA forms smaller, more achievable tasks with higher success rates. The delegation of responsibilities also allows for individual stages to be retried in event of generation failure, rather than all stages.


\section{Limitations}
While AURA automates reward design, domain randomization, and curriculum-driven policy training, a nontrivial amount of human effort is still required. Setting up the MuJoCo simulation environment, defining observation spaces, action spaces, and environment state variables, remains a manual process that can impact policy success. Similarly, deploying the final policy to real hardware requires careful integration and validation.

\textit{Human Evaluation:}
Although AURA’s training pipeline generates curricula autonomously, it still relies on human feedback at the end of each iteration to provide qualitative assessment of the deployed policy. While naively applying Vision-Language Models often fails to accurately describe robot data \cite{robo2vlm, robovqa}, recent frameworks demonstrate promising methods for automatically analyzing robot trajectories and delivering feedback \cite{autoeval, rewind}. These could be used in future work to fully automate feedback between iterations.

\textit{Exploration and VDB Content Bias:}
While AURA leverages its VDB to improve training reliability, this dependence introduces potential bias. If the database lacks diversity or contains inaccurate human feedback, subsequent retrieval-augmented generations may reinforce suboptimal patterns. AURA’s curriculum search process has the potential to converge prematurely on a narrow set of solutions. Future work could incorporate explicit exploratory sampling of curricula designs or adversarial selection of prior examples to counteract this VDB content bias.

\section{Conclusion}
This paper introduces \textbf{AURA (Autonomous Upskilling with Retrieval-Augmented Agents)}, a schema-centric curriculum RL framework leveraging multi-agent LLM architectures and RAG. By abstracting complex GPU-accelerated training pipelines into schema-validated YAML workflows, AURA reliably automates the design and refinement of multi-stage curricula, enabling policy training with iterative improvement.
Empirical evaluations demonstrate that AURA outperforms baseline methods, achieving a higher generation success rate and superior policy performance across locomotion and manipulation tasks, and effective zero-shot hardware deployment.
Real-world validation on a custom humanoid robot highlights AURA’s capability to autonomously produce highly robust policies directly from user-defined prompts. The resulting controller demonstrated stable outdoor locomotion, gait tracking, and disturbance rejection, including recovery from lateral perturbations and vertical drops. Overall, AURA advances the development of scalable, generalizable prompt-to-policy frameworks, demonstrating the ability to substantially reduce the manual effort required in reward design while generating complex, high-performing curricula.

\renewcommand*{\bibfont}{\footnotesize}
\printbibliography




\onecolumn

\appendix
\subsection{Implementation and Testing Details}
\subsubsection{Training via Proximal Policy Optimization (PPO)}  
We train the control policy with PPO, which constrains the policy update to prevent destructive parameter jumps.  
Let  
\[
r_t(\boldsymbol{\theta}) \;=\; 
\frac{\pi_{\boldsymbol{\theta}}\!\bigl(a_t \mid s_t\bigr)}
     {\pi_{\boldsymbol{\theta}_{\text{old}}}\!\bigl(a_t \mid s_t\bigr)}
\]
denote the probability ratio between the new and old policies, and let $A_t$ be the generalized advantage estimate at timestep $t$.  
The clipped surrogate objective is  
\[
L^{\text{CLIP}}(\boldsymbol{\theta}) \;=\;
\mathbb{E}_t\!\left[
  \min\!\Bigl(
      r_t(\boldsymbol{\theta})\,A_t,\;
      \text{clip}\!\bigl(r_t(\boldsymbol{\theta}),\,1-\epsilon,\,1+\epsilon\bigr)\,A_t
  \Bigr)
\right],
\]
where $\epsilon$ is the clipping parameter (we use $\epsilon=0.2$ unless otherwise noted).  
During training, we maximize $L^{\text{CLIP}}$ with Adam, using entropy regularization for exploration and a value–function loss with coefficient $c_v$.

\subsubsection{Score Calculation}
\label{sec:score_calc}

During one evaluation run, we launch \(N=1024\) parallel environments and roll each for
a horizon of \(T_{\max}=3000\) simulation steps (or until failure).  All three
metrics are normalized to \([0,1]\) for direct comparability.

\paragraph{Survival Score.}
Let \(T_i\in[1,T_{\max}]\) denote the number of steps survived by environment
\(i\).  The Survival Score is the mean fractional episode length
\[
\mathcal{S}_{\text{surv}}
\;=\;
\frac{1}{N}\sum_{i=1}^N\frac{T_i}{T_{\max}} .
\]

\paragraph{Linear‑Velocity Tracking Score.}
For step \(t\) in environment \(i\) let
\(\mathbf{v}^{\mathrm{cmd}}_{t,i}\in\mathbb{R}^2\) be the commanded planar
velocity and
\(\mathbf{v}^{\mathrm{loc}}_{t,i}\in\mathbb{R}^2\) the robot’s actual planar
COM velocity expressed in the local frame.  Define the squared tracking error
\[
e_{t,i} \;=\;
\bigl\lVert \mathbf{v}^{\mathrm{cmd}}_{t,i}-\mathbf{v}^{\mathrm{loc}}_{t,i}
\bigr\rVert_2^{\,2}.
\]
The YAML \texttt{exponential\_decay} entry with
\(\sigma=0.1\) corresponds to the per‑step reward
\[
r^{\text{lin}}_{t,i}
\;=\;
\exp\!\bigl(-e_{t,i}/(2\sigma^{2})\bigr),
\qquad \sigma=0.1 .
\]
Aggregating over time and averaging across the batch yields the normalized
Linear‑Velocity Tracking Score
\[
\mathcal{S}_{\text{lin}}
\;=\;
\frac{1}{N}\sum_{i=1}^{N}\frac{1}{T_{\max}}
\sum_{t=1}^{T_i} r^{\text{lin}}_{t,i}.
\]


\paragraph{Summary.}
The triplet
\(\bigl(\mathcal{S}_{\text{surv}},\mathcal{S}_{\text{lin}},\mathcal{S}_{\text{air}}\bigr)\)
captures robustness (survival), command‑following fidelity, and gait
coordination, respectively, and forms the basis of all quantitative comparisons
in the main paper.

\subsubsection{AURA Task Prompts for Training Launch}
\label{appendix/tasks}
These five user prompts were used as task inputs to AURA for evaluating its curriculum training-launch success rate across the custom humanoid and Berkeley Humanoid embodiments. For baseline comparisons, we used the prompts provided in each baseline’s open-source repository to generate their respective training files.

\begin{lstlisting}[language={},caption={Robust Bipedal Walking with Perturbation Resilience}]
I want to use a staged approach to train a humanoid robot for advanced locomotion with external perturbations. 
I want to deploy this policy onto hardware and walk outside, and I want the steps to be even (between the right and left legs) and smooth.
I want the walking to be very robust to both uneven terrain and perturbations.  
I want the training to have equal capabilities for forward and backwards walking, with an absolute max speed being 0.5m/s and max yaw being 0.5rad/s. 
You must generate AT LEAST 2 stages for this task.
\end{lstlisting}

\begin{lstlisting}[language={},caption={Terrain-Adaptive Walking}]
I want to use a multi-stage curriculum to train a humanoid robot to walk over irregular terrain, such as small rocks and low barriers.
The robot must learn to lift its feet high enough to avoid tripping and maintain a steady gait while stepping over obstacles of varying height (up to 0.02m).
The policy should be deployable outdoors and remain balanced when landing on slightly angled or unstable surfaces.
I want both forward and backward walking to be supported, with even step timing and foot clearance. 
You must generate AT LEAST 2 stages for this task.
\end{lstlisting}

\begin{lstlisting}[language={},caption={Precision Jumping Between Platforms}]
I want to use a staged approach to train a humanoid to jump onto and off of elevated platforms.
The policy should support both single jumps (from ground to platform) and double-jumps (from one platform to another).
The target jump height is 0.05m with target air time of 0.5s.
I want landing posture and knee angle to remain stable, and I want the robot to absorb impacts smoothly.
The final policy should transfer to hardware and be tested over rigid and slightly deformable platforms.
You must generate AT LEAST 2 stages for this task.
\end{lstlisting}

\begin{lstlisting}[language={},caption={Rhythmic Forward-Backward Hopping}]
I want to train a humanoid robot to perform continuous forward-backward hopping in a rhythmic, energy-efficient manner.
The hops should alternate directions every few steps and maintain even left-right force distribution.
I want the robot to be robust to mild perturbations during flight and landing.
Deployment should be feasible on a physical robot outdoors, with stability maintained on moderately uneven terrain.
Hop height should range between 0.05-0.1 meters with a frequency of ~1.5 Hz.
You must generate AT LEAST 2 stages for this task.
\end{lstlisting}

\begin{lstlisting}[language={},caption={Stable Lateral Walking with Perturbation Handling}]
I want to use a staged curriculum to train a humanoid robot to perform lateral (sideways) walking in both directions.
The walking should be smooth and balanced, with equal step distances between the left and right legs.
The policy should be robust to minor terrain irregularities and moderate lateral perturbations.
Maximum lateral velocity should be capped at +-0.3 m/s, and yaw rotation should be minimized during side-stepping.
I want the final policy to be deployable on hardware and capable of sustained lateral walking in outdoor environments.
You must generate AT LEAST 2 stages for this task.
\end{lstlisting}

\subsubsection{Training-Launch Comparison Details}
In the following section, we provide a more detailed description of the environments used to evaluate the CurricuLLM and Eureka training-launch success rates. All environments were evaluated over $100$ training launches.

\paragraph{CurricuLLM}
CurricuLLM's training-launch success rate was evaluated on their Berkeley Humanoid environment from \cite{ryu2024curricullm}, with all environment scripts and tasks taken directly from their open-source repository.

\paragraph{Eureka}
Eureka's training-launch success rate was evaluated on their only legged robot environment, ANYmal, across $100$ training-launch attempts, successfully launching training $12$ times. We also tested training-launch success across all embodiments provided in their open-source repository, which include the Gymnasium environments Shadow Hand, Franka Emika Panda, Ant, Humanoid, and Cartpole, as well as other robot environments such as ANYmal, Allegro, Ball Balance, and Quadcopter.

\subsection{AURA LLM Agent Prompt Templates}
\label{appendix/prompts}
The prompts below define the instruction templates used by AURA's LLM agents throughout the curriculum generation and training process. These templates structure how input data, such as user task prompts, stage descriptions, and past training artifacts, are transformed into schema-compliant outputs like workflows, configuration files, and reward functions. Placeholders in the form of \texttt{<INSERT\_...>} denote dynamic inputs that are automatically populated by the AURA framework at runtime with the appropriate content, ensuring that each prompt remains generalizable while retaining full contextual relevance for the target task.

\subsubsection{Curriculum LLM}
\mbox{}\\[-3ex]
\begin{lstlisting}[language={},]
You are an expert in reinforcement learning, CI/CD pipelines, GitHub Actions workflows, and curriculum design. Your task is to generate a complete curriculum with clearly defined steps for a multi-stage (staged curriculum learning) training process for a humanoid robot in simulation.
**MOST IMPORTANT DIRECTIONS:**
**<INSERT_TASK_PROMPT_HERE>**
Determine the number of stages based on the complexity of the task described.

You are to generate as follows:
1. A GitHub Actions workflow file that orchestrates the entire staged training process.
2. For each training stage that appears in the workflow, a separate text file containing in-depth, rich details that describe the objectives, expected changes to reward files, configuration file modifications, and overall curricular rationale.

-------------------------------------------------
**Important Guidelines:**
- IMPORTANT! The example baseline content parameters comes from a vector database of a past training workflow that's similar to the current task. Update them to fit the user's task more.
- IMPORTANT! Here is the EXPERT evaluation of the example training workflow results: <INSERT_EVALUATION_HERE>.
  - Use this evaluations by the EXPERT to update the parameters.
- **Workflow File Generation:**  
  Generate a workflow file named `generated_workflow.yaml` that orchestrates the staged training process. This file must:
  - Begin with the baseline structure provided via the placeholder
   `<INSERT_WORKFLOW_YAML_HERE>` and then be modified to support multiple training stages with explicit jobs for training and feedback (if feedback is needed).
  - Use the example reward file (provided via `<INSERT_REWARD_YAML_HERE>`) and example config file (provided via `<INSERT_CONFIG_YAML_HERE>`) as context.
  - Use the example randomize file (provided via `<INSERT_RANDOMIZE_YAML_HERE>`) as context to understand the domain randomization parameters for the scene.
  - Remember the workflow should use the same generated config, reward, and randomize that it defines in the stage description for training.

- **Detailed Stage Description Files:**  
  In addition to the workflow file, for every stage that is generated in the workflow, you must output a corresponding text file (named `generated_stage{X}_details.txt`, where `{X}` is the stage number) that contains very rich, in-depth details. Each of these text files must include:
  - A clear description of the training objective for that stage.
  - Precise descriptions of modifications to the config environment disturbances.
  - Explicit details on the expected modifications to the reward file.
  - A precise explanation of the configuration changes.
  - A discussion on the terrain/environment context (e.g., flat_scene, mixed_scene, height_scene).
  - Which gate to use (walk or jump)
  - How this stage fits into the overall curriculum progression 
  - Specify if the stage is starting from new or resuming from a checkpoint

-------------------------------------------------
**Additional Context:**
- The robot and training environment context is provided via <INSERT_ROBOT_DESCRIPTION_HERE>
- The generated workflow file must include explicit, detailed inline comments.
- IMPORTANT! The number of stages in the workflow and the stage descriptions generated should match exactly
-------------------------------------------------
**Output Format:**
**THIS IS THE MOST IMPORTANT PART! YOU MUST FOLLOW THESE DIRECTIONS EXACTLY! Return your output as separate YAML blocks (and nothing else) in the following format:**

file_name: "generated_workflow.yaml"
file_path: "../workflows/generated_workflow.yaml"
content: |
  # [The updated workflow.yaml content here with inline comments]

file_name: "generated_stage1_details.txt"
file_path: "../prompts/tmp/generated_stage1_details.txt"
content: |
  # [The in-depth description for Stage 1 with explicit detail on training objectives, reward configuration adjustments, and other parameters]
\end{lstlisting}

\subsubsection{RAG}
These prompts are used within the RAG Block to query, retrieve, and select suitable past workflows and training files to be used as context for the current task.

\paragraph{VDB Query LLM}
\mbox{}\\[-3ex]
\begin{lstlisting}
You are an expert in reinforcement learning for robotics. You are given a high-level task description related to training a robot policy.

Your job is to generate a short and precise **natural language query** that summarizes the core attributes of the task. This query will be used to retrieve similar configurations, rewards, and workflows from a vector database.

Your output should:
- Be a **single sentence**.
- Focus on **task type**, **terrain**, **number of curriculum stages**, and any **specific tuning goals** (e.g., push resistance, fine-tuning, checkpoint reuse).
- Avoid code, markdown, or explanations.

Task Description:
<INSERT_TASK_PROMPT_HERE>

Output:
(One-line query only)

\end{lstlisting}

\paragraph{Selector LLM}
\mbox{}\\[-3ex]
\begin{lstlisting}
You are an expert in reinforcement learning and robot curriculum design.

You are provided with:
- A high-level training task.
- A collection of YAML files from various past runs (each associated with a run ID). These include multiple stages of `reward`, `config`, and `randomize` YAMLs, as well as one or more `workflow` YAMLs per run.

Your job is to **select only the files needed** to best support the new multi-stage workflow generation. You should pick:
- Exactly one `workflow.yaml` file.
- One `reward.yaml`, `config.yaml`, and `randomize.yaml` file for **each** stage of the task (e.g., stage 1-3).

Output a JSON object mapping descriptive keys (e.g., `workflow`, `reward_stage1`, etc.) to the exact filenames you want to keep (from the file list provided). All other files will be deleted.

------------------------
## TASK DESCRIPTION
**<INSERT_TASK_PROMPT_HERE>**
------------------------
## HERE ARE THE EVALUATIONS FOR THE CANDIDATE RUNS:##
<INSERT_EVALUATIONS_HERE>
**Pick the runs whose evaluations look the most promising for the task.**
------------------------

## AVAILABLE FILES (filename -> truncated preview)
<INSERT_EXAMPLES_HERE>
------------------------

Your output must be in the following format:

```json
{
  "workflow": "run123_workflow.yaml",
  "reward_stage1": "run123_reward_stage1.yaml",
  "config_stage1": "run123_config_stage1.yaml",
  "randomize_stage1": "run123_randomize_stage1.yaml",
}

**If the run you are choosing has more stages, also put them into the output json block.**

Do NOT include any other text outside the fenced JSON block.


\end{lstlisting}

\subsubsection{Stage-Level LLM}
\mbox{}\\[-3ex]
\begin{lstlisting}
You are an expert in reinforcement learning, CI/CD pipelines, GitHub Actions workflows, and curriculum design. Your task is to generate the configuration files for a specific stage of a multi-stage (staged curriculum learning) training process for a humanoid robot in simulation. For this prompt, you are generating files for Stage {X} (for example, Stage 1, Stage 2, etc.). Use the inline comments from the workflow file (provided in <INSERT_WORKFLOW_YAML_HERE>) for Stage {X} to guide your modifications.

<INSERT_TASK_PROMPT_HERE>
-------------------------------------------------
**THESE ARE THE MOST IMPORTANT DIRECTIONS TO FOLLOW! Stage {X} Description:**
<INSERT_STAGE_DESCRIPTION_HERE>
-------------------------------------------------

**All reward, config, and randomize example files are selected from a vector database. Each of the examples have been selected by a higher-level LLM due to their fitting parameters for the task.**
**If you believe the provided files from the database are sufficiently good for the training of the given stage, you may choose to use the same parameters**

For the reward file:
- Use the content from <INSERT_REWARD_YAML_HERE> as a starting point.
- **Preserve all reward function keys** from the baseline. You must include every reward term from the original file in this stage's reward file.
- You may adjust the scalar weight (value) of each reward term to suit the stage. For keys that are not relevant at this stage, it is acceptable to set their values to 0.0 (effectively disabling them) while preserving the structure.
- If additional reward terms are necessary to support the stage goal, you may add new ones. New terms must use only the allowed function types listed below.
- **Allowed Functions (only use these functions when building reward expressions):**

<INSERT_SCHEMA_REWARD_EXPRESSIONS>

The top-level key in the reward file must be `reward:`.
- Add inline comments next to any changes, new reward terms, or adjustments explaining how they support the stage goal.

**Context of variables provided for reward functions based on the environment:**
<INSERT_REWARD_VARS_HERE>
- **Only the variables defined here can be used in the reward function calculations!**

**Here is an in depth example and explanation for creating reward functions:**
<INSERT_REWARD_EXAMPLE_HERE>
- These examples show the format of how to build our yaml based reward functions, and its equivalent in python. You have to use this style closely for new reward generation.

Notes:
- Input values to a reward function when setting a variable (e.g. lift_thresh: "0.2"), has to be a string. For example, in the lift_thresh case, even though it is setting lift_thresh
  to be a float 0.2, you still need to set it with a string.
- MAKE SURE to understand the variables, their types, and their shape for using them in reward functions.
    - IN PARTICULAR, double check for vectors the shapes much so there is no error!
- **IMPORTANT! REMEMBER! THIS IS JAX! You must use jax functions, conditions, arrays, etc for everything you generate in the reward function.**
    - Expressions like "vector: "rot_up - [0, 0, 1]" are not valid, you must use "rot_up - jp.array([0.0, 0.0, 1.0])"
    - USE VECTOR OPERATIONS!: In jax, you cannot use conditional statements like and, or, you have to use jax operations such as jp.where, or use bitwise such as &, |, etc.
- **You do not need to calculate a total reward! Just define individual reward functions, aggregating the rewards is handled elsewhere.**

-------------------------------------------------
For the configuration file:
- **Make sure to follow the exact structure of the config file, including the top-level keys. One of the top level classes is `environment:`, don't forget it!**
- Use the current baseline structure from <INSERT_CONFIG_YAML_HERE> as a reference for the structure and expected parameters.
- **Keep the structure of the config file.**
- **Do not add or remove any parameters**; only adjust the values to support the staged curriculum learning process.
- Adjust trainer parameters (e.g., learning rate, batch size, etc.) to support training stability as the curriculum advances.
- **resume_from_checkpoint:** This should be set to false if a new staged training is being done. If continueing from a previous checkpoint, this flag should be set to true.
  - refer to the stage description for which option to choose.
- Ensure that `batch_size` and `num_envs` remain powers of 2.
- The network parameters must remain consistent across all stages.
- Choose `scene_file` from the options (flat_scene, mixed_scene, and height_scene) according to the stage difficulty.
- Add inline comments next to any parameter changes explaining your reasoning.
- The `randomize_config_path` should be the corresponding generated randomize yaml file path for each stage.
-------------------------------------------------
**Baseline Randomization File for Context:**
<INSERT_RANDOMIZE_YAML_HERE>
-------------------------------------------------
**Instructions for the randomize.yaml file:**
- **Preserve the overall file structure and all keys from the baseline randomization file.**
- The randomize file contains parameters for domain randomization, including:
  - **geom_friction:** Randomizes friction properties for all geometry objects.
  - **actuator_gainprm:** Randomizes actuator gain parameters.
  - **actuator_biasprm:** Randomizes actuator bias parameters.
  - **body_ipos:** Randomizes body initial positions (center of mass positions).
  - **geom_pos:** Randomizes the positions of specific geometries.
  - **body_mass:** Randomizes the body mass (scaling factors or additional offsets).
  - **hfield_data:** Randomizes the heightfield (terrain) data values.
- Read the inline comments for each parameters in the randomize.yaml example for details on what the parameters do.
- You may adjust the numeric parameter values (e.g., the min and max values in the uniform distributions) to better support the training stage's objectives while keeping the structure intact.
- For parameters that are modified, add inline comments next to the changed values explaining how the adjustments support the stage goals (e.g., improved robustness to disturbances, adapting friction to better simulate challenging terrain, etc.).
- The top-level key of the file must remain "randomization:".
- **The randomization file path must match the one specified in the config yaml.**
-------------------------------------------------
**Output Format:**
**THIS IS THE MOST IMPORTANT PART! YOU MUST FOLLOW THESE DIRECTIONS EXACTLY! Return your output as separate YAML blocks (and nothing else) in the following format:**

file_name: "generated_reward_stage{X}.yaml"
file_path: "../rewards/generated_reward_stage{X}.yaml"
content: |
  # [The complete reward file for Stage {X} with inline comments]

file_name: "generated_config_stage{X}.yaml"
file_path: "../configs/generated_config_stage{X}.yaml"
content: |
  # [The complete configuration file for Stage {X} with inline comments]

file_name: "generated_randomize_stage{X}.yaml"
file_path: "../randomize/generated_randomize_stage{X}.yaml"
content: |
  # [The complete randomization file for Stage {X} with inline comments]

Do not output any text or explanation outside these YAML blocks.
-------------------------------------------------



\end{lstlisting}

\subsubsection{Feedback LLM}
\mbox{}\\[-3ex]
\begin{lstlisting}
Feedback integration for curriculum learning:
This is currently the Stage {current_stage} training feedback process based on the workflow.
Based on the previous training step's reward metrics and configurations, update the reward and configuration files for improved performance.
The metrics either 1) give information about the training process (e.g. entropy loss, policy loss, etc.), or
2) give the value each reward term contributes to the overall reward; the name of the eval/* for the reward matches exactly with that of the reward term in the reward file's functions.
Look for specific metrics, such as the loss and penalty values, to guide the updates (e.g. if loss numbers are too volatile, reduce the learning rate).

\end{lstlisting}
The above feedback prompt, together with the next stage’s description and all the stage’s metrics log file, is prepended to the Per-Stage LLM prompt for use by the Feedback LLM.


\subsection{AURA Outputs}
\paragraph{Configuration YAML}
\mbox{}\\[-3ex]
\begin{lstlisting}[language=yaml]
environment:
  scene_file: "../../../og_bruce/flat_scene.xml"  # Always flat terrain for this curriculum
  reward_config_path: "../rewards/generated_reward_stage1.yaml"  # Path to Stage 1 reward file
  obs_noise: 0.03  # Increased observation noise for robust sensorimotor learning
  imu_disturbs: true  # Enable IMU disturbances to enforce robust policy learning
  init_rand: true  # Extensive randomization in joint angles, body position, and mass
  big_min_kick_vel: 0.05  # Enable full big kick perturbations for robustness
  big_max_kick_vel: 0.12  # Enable full big kick perturbations for robustness
  big_kick_interval: 80  # Randomized interval for big kicks (see randomization file for full range)
  small_min_kick_vel: 0.01  # Enable small kicks for continuous disturbance
  small_max_kick_vel: 0.03  # Enable small kicks for continuous disturbance
  small_kick_interval: 10  # Randomized interval for small kicks (see randomization file for full range)
  fixed_command: false  # Commands randomized for full velocity/yaw tracking
  command_lin_vel_x_range: [-0.5, 0.5]  # Full forward/backward walking speed range
  command_lin_vel_y_range: [0.0, 0.0]  # No lateral walking required
  command_ang_vel_yaw_range: [-0.5, 0.5]  # Full yaw tracking range
  command_stand_prob: 0.15  # Robot is rarely asked to stand still; mostly active walking
  cutoff_freq: 3.0
  deadband_size: 0.01
  low_cmd_boost_scale: 1.5
  gait_frequency: [2.0, 2.25]  # Less range for better convergence and tracking
  gaits: ["walk"]  # Use walk gait for consistency
  foot_height_range: [0.03, 0.05]  # Smaller swing heights for better conergence of feet phase tracking

render:
  resolution: [640, 480]
  randomize_seed: 25
  n_steps: 3000
  view: "track_com"
  fps: null
  plot_actions: true
  plot_observations: true
  plot_rewards: true
  static_actions:
    10: 0
    11: 0
    12: 0
    13: 0
    14: 0
    15: 0

trainer:
  num_timesteps: 400_000_000  # Full stage budget: all training in a single phase (max allowed)
  num_evals: 13  # Evaluations every ~30M steps for robust progress monitoring
  reward_scaling: 1
  episode_length: 3000  # Longer episodes for sustained walking under disturbance
  normalize_observations: true
  action_repeat: 1
  unroll_length: 20
  num_minibatches: 32
  num_updates_per_batch: 4
  discounting: 0.97  # Higher discount for long-term stability and recovery
  learning_rate: 8.0e-5  # High learning rate for fast convergence
  entropy_cost: 5.0e-3  # Moderate entropy to encourage exploration
  num_envs: 8192
  batch_size: 512
  seed: 7  # New random seed for diversity
  clipping_epsilon: 0.18 # Slightly decreased for smaller updates in high-variance training

randomization:
  randomize: true
  randomize_config_path: "../randomize/generated_randomize_stage1.yaml"  # Path to Stage 1 randomization file

artifact:
  resume_from_checkpoint: false  # Training starts from scratch for single-stage curriculum
  master_path:
    abs_parent_path: null
    folder_prefix: "tmp/run"
    folder_index: -1
  checkpoint:
    checkpoint_dir: "model"
    folder_index: -1
    policy_params_fn: null

ppo_network:
  policy_hidden_layer_sizes: [512, 256, 128]  # Keep network structure consistent
  value_hidden_layer_sizes: [512, 256, 128]  # Keep network structure consistent
  activation: "swish"
  policy_obs_key: "state"
  value_obs_key: "privileged_state"
\end{lstlisting}

\paragraph{Reward YAML}
\mbox{}\\[-3ex]
\begin{lstlisting}[language=yaml]
reward:
  tracking_lin_vel:
    inputs:
      command_xy: "command[0:2]"
      local_vel_xy: "local_vel[0:2]"
    evaluations:
      - type: "sum_square"
        parameters:
          vector: "command_xy-local_vel_xy"
        output: "error"
      - type: "exponential_decay"
        parameters:
          error: "error"
          sigma: 0.09  # Tighter sigma for precise bidirectional velocity tracking; enforces true walking/stepping
    combination:
      type: "last"
    scale: 1.5  # Full reward for tracking commanded velocity (serves as survival reward as well)
    default_reward: 0.0

  tracking_ang_vel:
    inputs:
      command_yaw: "command[2]"
      base_ang_vel: "base_ang_vel"
    evaluations:
      - type: "norm_L2"
        parameters:
          vector: "command_yaw-base_ang_vel"
        output: "error"
      - type: "exponential_decay"
        parameters:
          error: "error"
          sigma: 0.1  # Tight sigma for accurate yaw tracking; prevents shuffling and enforces turning
    combination:
      type: "last"
    scale: 1.0  # Full reward for yaw tracking
    default_reward: 0.0

  lin_vel_z:
    inputs:
      vel_z: "xd.vel[0, 2]"
    evaluations:
      - type: "quadratic"
        parameters:
          value: "vel_z"
          weight: 1.0
    scale: 0.0  # Penalize vertical velocity not used
    default_reward: 0.0

  ang_vel_xy:
    inputs:
      ang_vel_xy: "xd.ang[0, :2]"
    evaluations:
      - type: "sum_square"
        parameters:
          vector: "ang_vel_xy"
    scale: -0.15  # Stronger penalty for roll/pitch instability under perturbations
    default_reward: 0.0

  orientation:
    inputs:
      rot_up_xy: "rot_up[0:2]"
    evaluations:
      - type: "sum_square"
        parameters:
          vector: "rot_up_xy"
    scale: -2.0  # Strong penalty for non-upright posture (enforces survival under heavy disturbance)
    default_reward: 0.0

  torques:
    inputs:
      torques: "qfrc_actuator"
    evaluations:
      - type: "norm_L2"
        parameters:
          vector: "torques"
      - type: "norm_L1"
        parameters:
          vector: "torques"
    combination:
      type: "sum"
    scale: -0.00004  # Slightly increased penalty for excessive actuator effort
    default_reward: 0.0

  action_rate:
    inputs:
      action: "action"
      last_act: "last_act"
    evaluations:
      - type: "sum_square"
        parameters:
          vector: "action - last_act"
    scale: -0.01  # Slight increase to encourage smoother actions under disturbance
    default_reward: 0.0

  feet_air_time:
    inputs:
      feet_air_time: "feet_air_time"
      first_foot_contact: "first_foot_contact"
      lift_thresh: "0.2"
      command_norm: "command_norm"
    evaluations:
      - type: "weighted_sum"
        parameters:
          values: "(feet_air_time - lift_thresh) * first_foot_contact"
          weights: 1.0
        output: "rew_air_time"
      - type: "binary"
        parameters:
          condition: "command_norm > 0.05"
          reward_value: "rew_air_time"
          else_value: 0.0
    scale: 2.0  # Enabled: rewards proper stepping and air time during walking
    default_reward: 0.0

  stand_still:
    inputs:
      commands_norm: "commands_norm"
      joint_angles: "joint_angles"
      default_pose: "default_pose"
    evaluations:
      - type: "norm_L1"
        parameters:
          vector: "joint_angles - default_pose"
        output: "norm_joint"
      - type: "binary"
        parameters:
          condition: "commands_norm < 0.01"
          reward_value: "norm_joint"
          else_value: 0.02
    scale: -0.35  # Enabled: penalizes deviation from rest pose only when robot is commanded to stand still
    default_reward: 0.0

  termination:
    inputs:
      done: "done"
      step: "step"
    evaluations:
      - type: "binary"
        parameters:
          condition: "done & (step < 500)"
          reward_value: -1.0
          else_value: 0.0
    scale: 1.0  # Maintain strong penalty for early episode termination/falling
    default_reward: 0.0

  foot_slip:
    inputs:
      contact: "foot_contact"
      foot_linear_velocity: "feet_site_linvel[:, 0:2]"
      foot_angular_velocity: "feet_site_angvel"
    evaluations:
      - type: "norm_L2"
        parameters:
          vector: "foot_linear_velocity"
        output: "linear_vel_norm"
      - type: "norm_L2"
        parameters:
          vector: "foot_angular_velocity"
        output: "angular_vel_norm"
    combination:
      type: "weighted_sum"
      parameters:
        vectors: ["linear_vel_norm * contact", "angular_vel_norm * contact"]
        weights: [1.0, 1.0]
    scale: -0.2  # Decreased penalty; enforces proper ground contact and reduces unnecessary foot motion
    default_reward: 0.0

  feet_phase:
    inputs:
      foot_z: "feet_pos[:, 2]"
      rz: "rz"
      first_foot_contact:  "first_foot_contact"
    evaluations:
      - type: "sum_square"
        parameters:
          vector: "(foot_z - rz)"
        output: "phase_err"
      - type: "exponential_decay"
        parameters:
          error: "phase_err"
          sigma: 0.001  # Ultra-tight sigma for phase adherence; enforces symmetric, phase-locked walking
    combination:
      type: "last"
    scale: 1.2  # More reward to encourge closer tracking of the feet phase
    default_reward: 0.0

  feet_clearance:
    inputs:
      foot_linear_velocity:   "feet_site_linvel"
      foot_z:                 "feet_site_pos[:,2]"
      max_foot_height:        "max_foot_height"
    evaluations:
      - type: "norm_L2"
        parameters:
          vector: "foot_linear_velocity[..., :2]"
        output: "vel_norm"
      - type: "absolute_difference"
        parameters:
          value1: "foot_z"
          value2: "max_foot_height"
        output: "delta_z"
      - type: "weighted_sum"
        parameters:
          values:  "delta_z * vel_norm"
          weights: 1.0
        output: "clearance_cost"
    combination:
      type: "last"
    scale: 0.02  # Enabled: encourages sufficient swing foot clearance, important for robust stepping
    default_reward: 0.0


\end{lstlisting}

\paragraph{Randomization YAML}
\mbox{}\\[-3ex]
\begin{lstlisting}[language=yaml]
randomization:
  geom_friction:
    - target: ALL
      distribution:
        uniform:
          minval: [0.6, 0.0, 0.0]
          maxval: [1.1, 0.0, 0.0]  # Broadened friction range for robustness to slip and diverse ground
      # Wider friction range supports learning stable walking under variable slip conditions

  actuator_kp_kd:
    - target: ALL
      distribution:
        uniform:
          minval: [-10, 0]
          maxval: [25, 0]  # Increased variability in actuation dynamics for robustness
      operation: add
      # Broader range encourages adaptability to actuator uncertainties

  body_ipos:
    - target: random_mass
      distribution:
        uniform:
          minval: [-0.03, -0.03, -0.03]
          maxval: [0.03, 0.03, 0.03]  # Greater initial posture diversity
      operation: add
      # Supports robustness to initialization errors and real-world pose variability

  geom_pos:
    - target: ['foot_contact_l', 'foot_contact_r']
      distribution:
        uniform:
          minval: [-0.004, -0.004, -0.004]
          maxval: [0.005, 0.004, 0.004]  # Increased foot geometry randomization
      operation: add
      # Prevents overfitting to a single contact configuration, improves generalization

  body_mass:
    - target: ALL
      distribution:
        uniform:
          minval: 0.9
          maxval: 1.1  # Full mass randomization for robustness
      operation: scale
    - target: random_mass
      distribution:
        uniform:
          minval: 0.0
          maxval: 1.0  # Full range for extra random mass
      # Full mass randomization ensures robustness to payload and hardware variation

  hfield_data:
    - target: ALL
      distribution:
        uniform:
          minval: 0
          maxval: 1  # No terrain randomization; always flat in this curriculum
      operation: scale
      # Flat surface only, no heightfield variation for this stage

\subsubsection{AURA Blind}
\paragraph{Configuration YAML}
\mbox{}\\[-3ex]
\begin{lstlisting}[language=yaml]
environment:
  scene_file: "../../../og_bruce/flat_scene.xml"        # Flat terrain only as specified for Stage 1
  reward_config_path: "../rewards/generated_reward_stage1.yaml"
  obs_noise: 0.03                                       # High observation noise for robustness
  imu_disturbs: true                                    # Enable IMU noise/disturbances
  init_rand: true  # Randomized initial pose for generalization
  big_min_kick_vel: 0.05  # Enable strong kicks from the start for robustness
  big_max_kick_vel: 0.2   # Increase max kick velocity for challenging perturbations
  big_kick_interval: 80   # More frequent big kicks
  small_min_kick_vel: 0.02  # Frequent, moderate small kicks
  small_max_kick_vel: 0.05  # Frequent, moderate small kicks
  small_kick_interval: 10   # Frequent small kicks for disturbance rejection
  fixed_command: false
  command_lin_vel_x_range: [-0.5, 0.5]                  # Full forward/backward command range for symmetric walking
  command_lin_vel_y_range: [0.0, 0.0]                   # No lateral walking for this stage
  command_ang_vel_yaw_range: [-0.5, 0.5]                # Full yaw command range for left/right turning
  command_stand_prob: 0.1                               # Small chance of stand command for stability
  cutoff_freq: 3.0
  deadband_size: 0.01
  gait_frequency: [2.0, 2.6]                           # Increased gait frequency for more natural/dynamic walking
  gaits: ["walk"]
  foot_height_range: [0.04, 0.08] # Larger step height range for generalization
  max_foot_height: 0.08

render:
  resolution: [640, 480]
  randomize_seed: 52
  n_steps: 3000
  view: "track_com"
  fps: null
  plot_actions: true
  plot_observations: true
  plot_rewards: true
  static_actions:
    10: 0
    11: 0
    12: 0
    13: 0
    14: 0
    15: 0

trainer:
  num_timesteps: 400_000_000                            # Single-stage, full allocation as per curriculum
  num_evals: 13                                         # ~30M steps per evaluation for efficiency
  reward_scaling: 1
  episode_length: 2000 # Sufficiently long to practice balance and command tracking for locomotion
  normalize_observations: true
  action_repeat: 1
  unroll_length: 20
  num_minibatches: 32
  num_updates_per_batch: 4
  discounting: 0.97
  learning_rate: 1.0e-4 # High learning rate for fast initial learning
  entropy_cost: 5.0e-3 # Slightly higher entropy for exploration at early stage
  num_envs: 8192
  batch_size: 512
  seed: 5
  clipping_epsilon: 0.2

randomization:
  randomize: true
  randomize_config_path: "../randomize/generated_randomize_stage1.yaml"

artifact:
  resume_from_checkpoint: false                         # Start from scratch for single-stage curriculum
  master_path:
    abs_parent_path: null
    folder_prefix: "tmp/run"
    folder_index: -1
  checkpoint:
    checkpoint_dir: "model"
    folder_index: -1
    policy_params_fn: null

ppo_network:
  policy_hidden_layer_sizes: [512, 256, 128]
  value_hidden_layer_sizes: [512, 256, 128]
  activation: "swish"
  policy_obs_key: "state"
  value_obs_key: "privileged_state"
    
\end{lstlisting}

\paragraph{Reward YAML}
\mbox{}\\[-3ex]
\begin{lstlisting}[language=yaml]
reward:
  tracking_xy_velocity:
    inputs:
      command_xy:   "command[0:2]"
      local_vel_xy: "local_vel[0:2]"
    evaluations:
      - type: "sum_square"
        parameters:
          vector: "command_xy-local_vel_xy"
        output: "error"
      - type: "exponential_decay"
        parameters:
          error: "error"
          sigma: 0.20   # Stricter velocity tracking (was 0.25); better command following for both forward/backward walking
    combination:
      type: "last"
    scale: 3.0
    default_reward: 0.0

  tracking_yaw:
    inputs:
      command_yaw: "command[2]"
      base_ang_vel: "base_ang_vel"
    evaluations:
      - type: "norm_L2"
        parameters:
          vector: "command_yaw-base_ang_vel"
        output: "error"
      - type: "exponential_decay"
        parameters:
          error: "error"
          sigma: 0.16  # Stricter yaw tracking (was 0.2); ensures accurate directional control
    combination:
      type: "last"
    scale: 2.0
    default_reward: 0.0

  feet_lift_time:
    inputs:
      feet_air_time:    "feet_air_time"
      first_foot_contact: "first_foot_contact"
      lift_thresh:      "0.2"
      command_norm:     "command_norm"
    evaluations:
      - type: "weighted_sum"
        parameters:
          values: "(feet_air_time - lift_thresh) * first_foot_contact"
          weights: 1.0
        output: "rew_air_time"
      - type: "binary"
        parameters:
          condition:   "command_norm > 0.05"
          reward_value: "rew_air_time"
          else_value:   0.0
    combination:
      type: "last"
    scale: 1.6
    default_reward: 0.0

  feet_swing_height:
    inputs:
      swing_peak:             "swing_peak"
      first_foot_contact:     "first_foot_contact"
      max_foot_height:        "max_foot_height"
    evaluations:
      - type: "sum_square"
        parameters:
          vector: "(swing_peak / max_foot_height - 1.0) * first_foot_contact"
        output: "height_err"
    combination:
      type: "last"
    scale: 0.0
    default_reward: 0.0

  symmetry:
    inputs:
      joint_angles: "joint_angles"
    evaluations:
      - type: "norm_L1"
        parameters:
          vector: "joint_angles[0:12] - joint_angles[12:24]"
    combination:
      type: "last"
    scale: -0.18      # Penalizes asymmetric joint movement; crucial for symmetric bidirectional walking
    default_reward: 0.0

  foot_scuff:
    inputs:
      contact: "first_site_contact"
      foot_linear_velocity: "feet_site_linvel"
      foot_angular_velocity: "feet_site_angvel"
    evaluations:
      - type: "norm_L2"
        parameters:
          vector: "foot_linear_velocity"
        output: "linear_vel_norm"
      - type: "norm_L2"
        parameters:
          vector: "foot_angular_velocity"
        output: "angular_vel_norm"
    combination:
      type: "weighted_sum"
      parameters:
        vectors: ["linear_vel_norm * contact", "angular_vel_norm * contact"]
        weights: [1.0, 1.0]
    scale: -0.018
    default_reward: 0.0

  linear_velocity_z:
    inputs:
      vel_z: "xd.vel[0, 2]"
    evaluations:
      - type: "quadratic"
        parameters:
          value: "vel_z"
          weight: 1.0
    scale: -0.08
    default_reward: 0.0

  angular_velocity_xy:
    inputs:
      ang_vel_xy: "xd.ang[0, :2]"
    evaluations:
      - type: "sum_square"
        parameters:
          vector: "ang_vel_xy"
    scale: -0.15
    default_reward: 0.0

  orientation_penalty:
    inputs:
      rot_up_xy: "rot_up[0:2]"
    evaluations:
      - type: "sum_square"
        parameters:
          vector: "rot_up_xy"
    scale: -3.0
    default_reward: 0.0

  penalty_torques:
    inputs:
      torques: "qfrc_actuator"
    evaluations:
      - type: "norm_L2"
        parameters:
          vector: "torques"
      - type: "norm_L1"
        parameters:
          vector: "torques"
    combination:
      type: "sum"
    scale: -0.0005
    default_reward: 0.0

  action_smoothness:
    inputs:
      action: "action"
      last_act: "last_act"
    evaluations:
      - type: "sum_square"
        parameters:
          vector: "action - last_act"
    scale: -0.07
    default_reward: 0.0

  low_command_stand:
    inputs:
      commands_norm: "commands_norm"
      joint_angles: "joint_angles"
      default_pose: "default_pose"
    evaluations:
      - type: "norm_L1"
        parameters:
          vector: "joint_angles - default_pose"
        output: "norm_joint"
      - type: "binary"
        parameters:
          condition: "commands_norm < 0.05"
          reward_value: "norm_joint"
          else_value: 0.02
    scale: -0.4
    default_reward: 0.0

  falling_penalty:
    inputs:
      done: "done"
      step: "step"
    evaluations:
      - type: "binary"
        parameters:
          condition: "done & (step < 1000)"
          reward_value: -1.0
          else_value: 0.0
    scale: 2.0
    default_reward: 0.0

  alive:
    inputs: {}
    evaluations:
      - type: "binary"
        parameters:
          condition: "True"
          reward_value: 1.0
          else_value: 0.0
    scale: 1.0
    default_reward: 0.0
\end{lstlisting}

\paragraph{Randomization YAML}
\mbox{}\\[-3ex]
\begin{lstlisting}[language=yaml]
randomization:
  geom_friction:
    - target: ALL
      distribution:
        uniform:
          minval: [0.4, 0, 0]
          maxval: [1.0, 0, 0]
      # Wide friction range for robust contact handling on flat terrain

  actuator_gainprm:
    - target: ALL
      distribution:
        uniform:
          minval: [-25, 0, 0, 0, 0, 0, 0, 0, 0, 0]
          maxval: [25, 0, 0, 0, 0, 0, 0, 0, 0, 0]
      operation: add
      # Wide gain range for actuator robustness (sim2real transfer)

  actuator_biasprm:
    - target: ALL
      distribution:
        uniform:
          minval: [0, 0, 0, 0, 0, 0, 0, 0, 0, 0]
          maxval: [0, 0, 0, 0, 0, 0, 0, 0, 0, 0]
      operation: add

  body_ipos:
    - target: random_mass
      distribution:
        uniform:
          minval: [-0.022, -0.022, -0.022]
          maxval: [0.022, 0.022, 0.022]
      operation: add
      # Increased range (+-0.022m) for robustness to random COM shifts, critical for perturbation rejection

  geom_pos:
    - target: ['foot_contact_l', 'foot_contact_r']
      distribution:
        uniform:
          minval: [-0.005, -0.005, -0.005]
          maxval: [0.005, 0.005, 0.005]
      operation: add
      # Increased range for geometry randomization, improves robustness to foot contact placement errors

  body_mass:
    - target: ALL
      distribution:
        uniform:
          minval: 0.9
          maxval: 1.1
      operation: scale
    - target: random_mass
      distribution:
        uniform:
          minval: 0.2
          maxval: 0.8

  hfield_data:
    - target: ALL
      distribution:
        uniform:
          minval: 0
          maxval: 1
      operation: scale
      # No terrain randomization; included for completeness, but flat_scene.xml disables heightfield effects
\end{lstlisting}

\end{document}